\documentclass[letterpaper, 10 pt, conference]{ieeeconf}  % Comment this line out if you need a4paper

\IEEEoverridecommandlockouts                              % This command is only needed if 
\usepackage[ruled,linesnumbered]{algorithm2e}

\newcommand{\cmark}{\checkmark}
\newcommand{\xmark}{$\times$}

\usepackage{booktabs}

\usepackage{makecell}
\usepackage{amsmath}
\usepackage{amssymb}
\usepackage{graphicx}
\usepackage{booktabs}
\usepackage{hyperref} 
\usepackage{xcolor}
\title{\LARGE \bf
Don't Freeze, Don't Crash: Extending the Safe Operating Range of Neural Navigation in Dense Crowds
}

\newif\ifanonymous
\ifanonymous
  \author{Anonymous Authors}
\else
  \author{Jiefu Zhang, Yang Xu, Vaneet Aggarwal%
  \thanks{The authors are with Purdue University, USA. Emails: \{zhan4018, xu1720, vaneet\}@purdue.edu}%
  }
\fi

\begin{document}

\maketitle
\IEEEpeerreviewmaketitle
\thispagestyle{empty}
\pagestyle{empty}

\begin{abstract}
Navigating safely through dense crowds requires collision avoidance that generalizes beyond the densities seen during training. Learning-based crowd navigation can break under out-of-distribution crowd sizes due to density-sensitive observation normalization and social-cost scaling, while analytical solvers often remain safe but freeze in tight interactions. We propose a reinforcement learning approach for dense, variable-density navigation that attains zero-shot density generalization using a density-invariant observation encoding with density-randomized training and physics-informed proxemic reward shaping with density-adaptive scaling. The encoding represents the distance-sorted $K$ nearest pedestrians plus bounded crowd summaries, keeping input statistics stable as crowd size grows.  Trained with $N\!\in\![11,16]$ pedestrians in a $3\mathrm{m}\times3\mathrm{m}$ arena and evaluated up to $N\!=\!21$ pedestrians ($1.3\times$ denser), our policy reaches the goal in $>99\%$ of episodes and achieves $86\%$ collision-free success in random crowds, with markedly less freezing than analytical methods and a $>\!60$-point collision-free margin over learning-based benchmark methods. Codes are available at \href{https://github.com/jznmsl/PSS-Social}{https://github.com/jznmsl/PSS-Social}.
\end{abstract}

\section{Introduction}
\label{sec:introduction}

Modern deep reinforcement learning (DRL) has advanced social robot navigation in indoor service environments~\cite{chen2019crowd,everett2018motion}, pedestrian-rich sidewalks~\cite{xie2023drlvo}, and structured corridors~\cite{liu2021decentralized} through better observation encodings~\cite{chen2019crowd,liu2021decentralized}, interaction modeling~\cite{chen2019crowd,di2024hyp}, and reward design~\cite{xie2023drlvo,chen2017socially}. Among the various social navigation scenarios, crowded navigation is uniquely challenging since the robot operates in environments with densely crowded moving pedestrians, while the crowd density varies unpredictably. For example, a hospital corridor that is nearly empty at night can reach up to $2$ pedestrians per square meter (ped/m$^2$) during a shift change, and a robot trained on moderate traffic may still be required to operate safely when an unexpected event increases local density. As a result, these robots will encounter interaction densities at deployment that exceed those seen during training. The resulting challenge is to navigate through dense crowds avoiding collisions, with zero-shot generalizations to unseen crowd densities. We specifically refer to densities larger than $1$ ped/m$^2$ as dense crowds.

Despite progress in DRL crowd navigation, the above challenge remains largely unaddressed. To our knowledge, prior work has not trained DRL policies under high interaction density and then systematically evaluated collision-free safety under out-of-distribution (OOD) density shifts. In sparser crowds, each moving pedestrian can be treated as a largely independent obstacle, allowing ample free space to plan around individual pedestrian agent sequentially. At densities above $ 1$~ped/m$^2$, the robot is simultaneously within a close range of multiple pedestrians, and the pedestrians within the close range may vary at each step. Since pedestrian agents are dynamic and constrain one another, this may result in coupled interactions where the avoidance of one agent displaces the robot into another agent's path.

Many prior DRL frameworks assume fixed-dimensional observation spaces \cite{everett2018motion,raffin2021stable}, which means the crowd state must be encoded into a fixed-length vector regardless of the present pedestrian numbers. This encoding inevitably involves inactive-slot padding when crowd size varies \cite{everett2018motion,chen2017socially}, and standard learned normalizers such as VecNormalize applied to such observations produce distributional artifacts when the number of active slots at test time differs from training. These effects are amplified in our dense, zero-shot density setting, where test-time crowds can exceed the training distribution and more neighbors become simultaneously relevant. While variable-length architectures such as attention pooling~\cite{chen2019crowd} and graph networks~\cite{liu2021decentralized,di2024hyp} can sidestep explicit padding, they introduce their own density-dependent challenges. Specifically, as crowd density increases, attention mechanisms suffer from weight dilution, where the influence of critical neighbors is washed out by the aggregation of numerous distant agents, and graph message-passing scales poorly, leading to oversmoothed feature representations.

Conversely, analytical approaches such as Optimal Reciprocal Collision Avoidance (ORCA)~\cite{van2011reciprocal} and Social Force Model (SFM)~\cite{helbing1995social} do not suffer from distribution shift but are fundamentally limited by their reliance on explicit geometric constraints and repulsive fields. In high-density scenarios, the set of feasible collision-free velocities often vanishes, forcing these solvers into the ``Freezing Robot Problem''~\cite{trautman2010unfreezing} where the robot minimizes risk by halting progress entirely. This behavior is overly conservative; while our experiments confirm that ORCA maintains high safety, it frequently succumbs to inefficient deadlocks, resulting in episodes with much higher freezing rates. Furthermore, these methods struggle to model the non-cooperative or noisy behaviors of real pedestrians that do not strictly adhere to reciprocity assumptions.

In this work, we address the above challenges and propose an efficient framework for robot navigation under dense crowds. To design a reliable observation encoder robust to density changes, we introduce a density-invariant encoding scheme that truncates to the $K$ nearest pedestrians and assigns them to distance-sorted slots, keeping the input dimension fixed while preserving consistent slot semantics as neighbors change.
To retain wider-crowd context without reintroducing density-dependent input size, we augment the neighbor list with  fixed and bounded crowd-summary scalars whose magnitudes remain comparable across densities, stabilizing standard normalization under OOD occupancy.
We pair this representation with density-randomized training by sampling $N$ uniformly within an interval, so the learned normalizer observes varying crowd sizes during training while remaining well-defined at test time.
To mitigate freezing, on the reward side, we model the dynamics between the robot and pedestrian agents as potential-based reward shaping~\cite{ng1999policy} grounded in Hall's interpersonal distance theory~\cite{hall1966hidden}, using an SFM-inspired repulsion potential~\cite{helbing1995social} as a proxemic reward. We further compute the intrinsic reward by applying density-adaptive scaling to make the intrinsic reward remain comparable throughout dense, coupled interactions. Together, these choices yield an algorithm that preserves ORCA-like safety without inheriting ORCA's tendency to freeze, while generalizing zero-shot beyond the training densities to maintain high collision-free success and near-universal goal reaching. The technical contributions of this paper are summarized as follows:
\begin{itemize}
\item We identify two structural failure modes that hinder zero-shot safety generalization across crowd densities. Learning-based methods suffer from observation distribution shifts and attention dilution that compromise safety, whereas analytical solvers succumb to freezing due to rigid geometric constraints.
\item We propose a density-invariant observation encoding via a distance-sorted $K$-nearest-neighbor truncation along with bounded crowd-summary features with fixed dimensions. Combining with density-randomized training, we enable standard observation encoding to remain stable under density shifts.
\item We introduce a novel reward shaping method that combines the potential-based shaping with density-adaptive scaling. Our ablation studies show that the mechanisms are individually insufficient but jointly necessary.
\item We study densely crowded navigation with zero-shot density generalization. Specifically, we evaluate zero-shot density generalization by training with $N\in[11,16]$ pedestrians in a $3\,\mathrm{m}\times3\,\mathrm{m}$ arena (1.22--1.78~ped/m$^2$) and testing up to $N{=}21$ (2.33~ped/m$^2$, $1.3\times$ the training maximum), achieving 86\% mean collision-free success with $>$99\% goal-reaching and a $>$60 percentage-point advantage over several existing algorithms under the same densification protocol.
\end{itemize}

\section{Related Work}\label{sec:related}

\subsection{Learning-based robot navigation in crowds}
DRL has been widely adopted for socially-aware robot navigation, often by modeling pedestrians as dynamic agents and learning a reactive policy in simulation.
Early approaches such as CADRL and its extensions learn collision avoidance behaviors in multi-agent settings using compact state representations and feedforward policies~\cite{chen2017socially,everett2018motion}.
To better capture interactions among multiple humans, later work introduced aggregation mechanisms that map a variable number of neighbors into a fixed-size embedding, including attention pooling in SARL-style methods~\cite{chen2019crowd} and structured interaction models such as DS-RNN~\cite{liu2021decentralized}.
Other lines incorporate geometric priors such as velocity-obstacle reasoning into the learning loop~\cite{xie2023drlvo} or use graph-based encoders and exploration objectives to improve planning in social settings~\cite{di2024hyp,pathak2017curiosity}. Emerging approaches also explore alternative learning paradigms for safe navigation, including diffusion-based generative planners with uncertainty-driven adaptive replanning for dynamic obstacle avoidance \cite{punyamoorty2025dynamic}, as well as vision-language-model scoring to shape socially appropriate motion online \cite{song2024vlm}.
Despite strong performance in standard benchmarks, a recurring limitation is that evaluations are often conducted at relatively low crowd densities and within training-like conditions, leaving open how these methods scale to denser, OOD deployments.

\begin{table}[t]
\caption{Comparison of DRL crowd navigation methods. Max density
refers to the highest density evaluated in the original work.}
\label{tab:comparison}
\centering
\setlength{\tabcolsep}{3pt}
\footnotesize
\begin{tabular}{@{}lccccc@{}}
\toprule
\textbf{Method}  & \makecell{\textbf{Max}\\\textbf{ped/m\textsuperscript{2}}} & \makecell{\textbf{High}\\\textbf{density}} & \makecell{\textbf{Safety}\\\textbf{metric}} & \makecell{\textbf{OOD}\\\textbf{density}} \\
\midrule
CADRL~\cite{everett2018motion}           & 0.10 & \xmark & \cmark & \xmark \\
SARL~\cite{chen2019crowd}               & 0.10 & \xmark & \cmark & \xmark \\
DenseCAvoid~\cite{sathyamoorthy2020densecavoid}  & {$>$1}\textsuperscript{$\dagger$} & \cmark\textsuperscript{$\dagger$} & \xmark & \xmark \\
CrowdSteer~\cite{liang2021crowd}    & {$>$1}\textsuperscript{$\dagger$} & \cmark\textsuperscript{$\dagger$} & \xmark & \xmark \\
Frozone~\cite{sathyamoorthy2020frozone}   & 0.10 & \xmark & \cmark & \xmark \\
DS-RNN~\cite{liu2021decentralized}     & 0.40 & \xmark & \cmark & \cmark \\
DRL-VO~\cite{xie2023drlvo}             & 0.22 & \xmark & \cmark & \cmark \\
SafeCrowdNav~\cite{xu2023safecrowdnav}& 0.10 & \xmark & \cmark & \xmark \\
Hyp2Nav~\cite{di2024hyp}       & 0.10 & \xmark & \cmark & \xmark \\
Sigal~\textit{et al.}~\cite{sigal2023improving}  & 0.10\textsuperscript{$\ddagger$} & \xmark & \xmark & \cmark \\
\midrule
\textbf{Ours}                         & \textbf{2.33} & \cmark & \cmark & \cmark \\
\bottomrule
\end{tabular}

\vspace{2pt}
{\scriptsize
\textsuperscript{$\dagger$}Real-world demo at noted density; controlled sim.\ density unreported or {$<$}0.5 ped/m\textsuperscript{2}.\\
\textsuperscript{$\ddagger$}Studies training generalization; evaluates at standard CrowdNav density.
}
\vspace{-8pt}
\end{table}

\subsection{Dense crowds, social compliance, and safety evaluation}
A subset of learning-based systems explicitly targets dense crowds by emphasizing real-time performance and anticipatory behaviors, including DenseCAvoid and CrowdSteer~\cite{sathyamoorthy2020densecavoid,liang2021crowd}.
Relatedly, Frozone focuses on reducing the freezing behavior that emerges in human crowds by shaping behaviors toward pedestrian-friendly motion~\cite{sathyamoorthy2020frozone}.
Complementary to algorithmic development, SafeCrowdNav highlights the need for systematic safety evaluation across diverse scenes and metrics~\cite{xu2023safecrowdnav}. In parallel, recent community efforts advocate standardized, repeatable benchmarks and metric suites for social robot navigation~\cite{francis2025principles}, and new datasets target socially-compliant navigation behaviors at scale~\cite{karnan2022socially}.
However, many dense-crowd demonstrations do not report controlled simulator densities or OOD density stress tests, which complicates direct comparisons and leaves open whether success transfers to densities beyond those seen during development.

\subsection{Generalization across crowd sizes and OOD densities}
Generalization remains a central challenge in crowd navigation because increasing the number of pedestrians changes interaction structure and can induce distribution shift.
DS-RNN and DRL-VO report evaluations beyond their nominal settings and include safety-oriented metrics~\cite{liu2021decentralized,xie2023drlvo}, while \cite{sigal2023improving} studies training regimes aimed at improving generalization in social navigation.
More broadly, safe reinforcement learning has been studied for navigation in dynamic environments~\cite{zhou2023safe}, but dense, tightly coupled crowds with explicit zero-shot density generalization requirements remain underexplored.

\subsection{Analytical collision avoidance and social models}
Analytical methods such as ORCA and social-force-based models offer strong geometric safety priors and avoid learning-induced distribution shift~\cite{van2011reciprocal,helbing1995social}.
Yet in dense interacting crowds, feasible collision-free solutions can shrink dramatically, leading to overly conservative behavior and the ``Freezing Robot Problem''~\cite{trautman2010unfreezing}.
In parallel, socially compliant navigation often draws on proxemics and human-interaction principles~\cite{hall1966hidden}, but translating these into policies that remain both safe and efficient at high densities, especially under OOD crowd sizes, is still challenging.

% ============================================================
% METHODS
% ============================================================

\section{Methods}
\label{sec:methods}

\begin{figure*}[t]
  \centering
  \includegraphics[width=\textwidth]{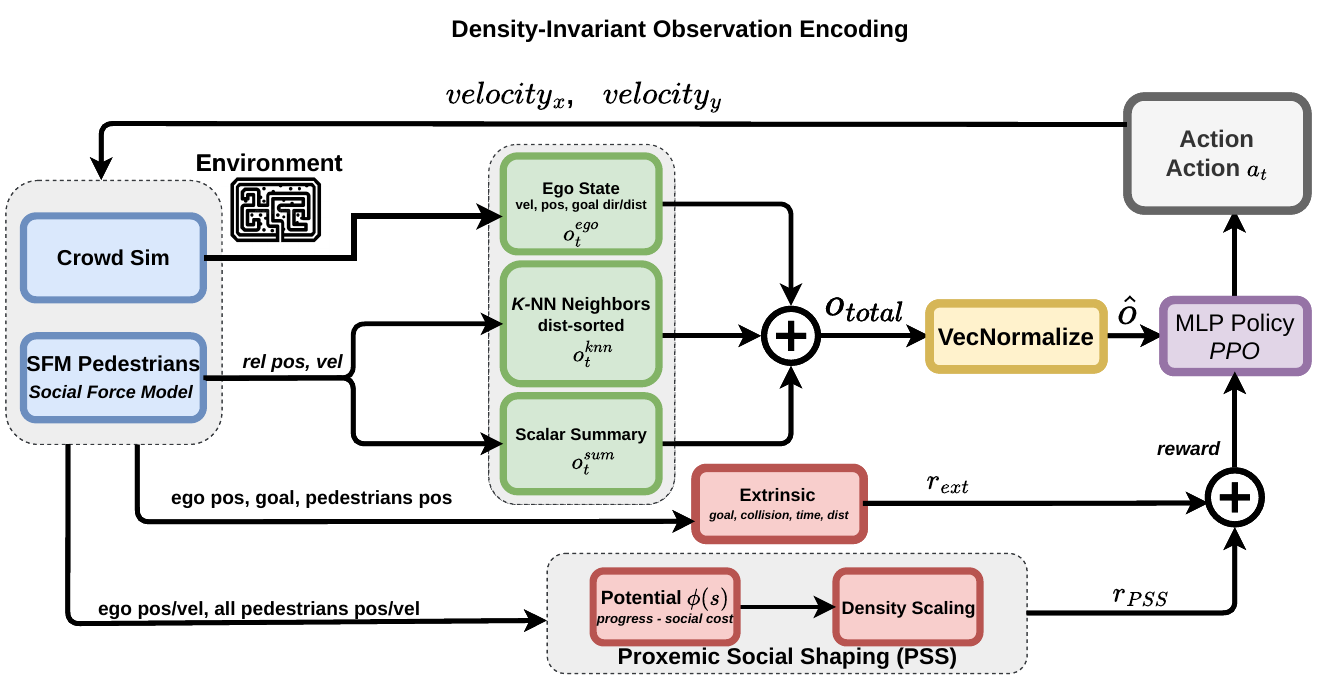}
  \caption{Overview of the PSS-Social pipeline. The simulator state is mapped to a fixed-dimensional observation consisting of ego-goal features, distance-sorted neighbor slots with $K$-cap truncation, and bounded crowd-summary scalars. The normalized observation is passed to an MLP policy trained with PPO. The training reward adds potential-based proxemic shaping with density-adaptive scaling to the environment’s extrinsic navigation reward.}
  \label{fig:pipeline}
\end{figure*}

We refer to the overall approach proposed in this section as \emph{PSS-Social}. The end-to-end pipeline is illustrated in Fig.~\ref{fig:pipeline}, and Algorithm~\ref{alg:pss_training} provides the complete training procedure.

\subsection{Problem Formulation}
We model social navigation as a partially observable Markov decision process (POMDP) with state $s_t\in\mathcal{S}$, observation $o_t\in\mathcal{O}$,
continuous action $a_t\in\mathcal{A}$, and transition dynamics $p(s_{t+1}| s_t,a_t)$.
The ego policy $\pi_\theta(a_t| o_t)$ is trained to maximize the discounted return
\begin{equation}
\theta^\star =\arg\max_{\theta}\;\mathbb{E}\!\left[\sum_{t=0}^{T-1}\gamma^t r_t\right],
\label{eq:rl_objective}
\end{equation}
with discount $\gamma\in(0,1)$. In our simulator, the ego action is a low-dimensional continuous control.

We consider a planar navigation domain $\Omega\subset\mathbb{R}^2$ with finite area $|\Omega|$.
At each time step, the environment contains one ego robot and $N$ pedestrians. We denote the ego state by $x_t^{0}=(p_t^{0},v_t^{0})\in\mathbb{R}^2\times\mathbb{R}^2$ containing both position and velocity, and the $i$-th pedestrian state by $x_t^{i}=(p_t^{i},v_t^{i})\in\mathbb{R}^2\times\mathbb{R}^2$ for $i\in\{1,\dots,N\}$.
The full POMDP state can be written as $s_t=\big(x_t^{0},x_t^{1},\dots,x_t^{N},g,\xi\big)\in\mathcal{S},$ where $g\in\Omega$ denotes the ego goal.
The variable $\xi$ summarizes additional episode parameters that influence transitions such as the scenario configuration and pedestrian goal parameters.

We characterize density by the episode-level average $\rho = \frac{N}{|\Omega|}$, and refer to settings with $\rho > 1$ ped/m$^2$ as dense. While $\rho$ provides a global measure, the instantaneous difficulty faced by the ego is governed by the
local interaction load, i.e., how many pedestrians fall within an interaction radius $r>0$:
\begin{equation}
n_t(r) = \sum_{i=1}^{N}\mathbb{I}\!\left[\|p_t^{i}-p_t^{0}\|\le r\right],
\label{eq:local_count}
\end{equation}
with $\mathbb{I}$ being the indicator function. Intuitively, $n_t(r)$ is large in dense regimes, meaning the ego repeatedly encounters simultaneous multi-agent interactions
rather than isolated pairwise encounters. For approximately uniform crowds, $\mathbb{E}[n_t(r)]\approx \rho\,\pi r^2$, so $\rho>1$ ped/m$^2$ implies multiple neighbors within typical interaction radii.

The transition kernel $p(s_{t+1} | s_t,a_t)$ is induced by the ego control $a_t$ and the interaction-coupled crowd dynamics.
In our simulator, pedestrians evolve according to a rule-based controller parameterized by the episode context $\xi$,
and we treat the overall dynamics as a black-box transition model for policy learning.

The policy receives an observation $o_t=\psi(s_t)$ rather than the full state. In our implementation, $\psi(\cdot)$ produces a fixed-dimensional vector composed of ego-and-goal features, relative measurements to nearby pedestrians ordered by proximity and truncated to a fixed budget, along with a small set of crowd-summary scalars.

\subsection{Density-Invariant Observation Encoding}
\label{subsec:obs}

To avoid density-induced representation drift, we design
$\psi(\cdot)$ to produce a fixed-dimensional observation with stable semantics as the crowd becomes denser. We construct
\begin{equation}
o_t = \psi(s_t) = \Big[o_t^{\mathrm{ego}}; o_t^{\mathrm{knn}};  o_t^{\mathrm{sum}}\Big],
\label{eq:obs_concat}
\end{equation}
where $o_t^{\mathrm{ego}}$ encodes ego-and-goal information, $o_t^{\mathrm{knn}}$ encodes a fixed budget of nearby pedestrians using proximity-ordered slots,
and $o_t^{\mathrm{sum}}$ is a small set of bounded crowd-summary scalars. Specifically, let $\Delta g_t=g-p_t^{0}$ and $d_g(t)=\|\Delta g_t\|$.
We use
\begin{equation}
o_t^{\mathrm{ego}}
=
\Big[v_t^{0}, p_t^{0}, \tfrac{\Delta g_t}{\max(d_g(t),\epsilon)}, d_g(t)\Big],
\label{eq:ego_obs}
\end{equation}
with $\epsilon>0$ for numerical stability. For each pedestrian $i$, define relative features
\begin{equation}
\Delta p_t^{i}=p_t^{i}-p_t^{0},\qquad \Delta v_t^{i}=v_t^{i}-v_t^{0},\qquad d_{t,i}=\|\Delta p_t^{i}\|.
\nonumber
\end{equation}
Let $\pi_t$ sort pedestrians by distance, i.e., $d_{t,\pi_t(1)}\le \dots \le d_{t,\pi_t(N)}$.
We use a fixed slot budget $K_{\max}$ (the maximum number of pedestrian slots represented) and an
active-neighbor cap $K_{\mathrm{cap}}\le K_{\max}$.
Define $K_t \triangleq \min(N, K_{\mathrm{cap}})$.
We allocate $K_{\max}$ slots and fill only the $K_t$ closest pedestrians:
\begin{equation}
o_t^{\mathrm{knn}}=\big[z_{t,1},\dots,z_{t,K_{\max}}\big],
\end{equation}
where each slot is
\begin{equation}
z_{t,k}=
\begin{cases}
\big[\mathrm{clip}(\Delta p_t^{\pi_t(k)}),\;\mathrm{clip}(\Delta v_t^{\pi_t(k)})\big], & k\le K_t,\\[2pt]
z_{\mathrm{pad}}, & k> K_t,
\end{cases}
\label{eq:knn_slots}
\end{equation}
$\mathrm{clip}(\cdot)$ bounds values to fixed ranges and $z_{\mathrm{pad}}$ is a constant ``far-away'' padding sentinel.
Distance ordering ensures each slot has a consistent meaning as the $k$-th closest pedestrian, stabilizing per-slot semantics as $N$ varies.
Crucially, keeping $K_{\mathrm{cap}}$ fixed during training ensures that slots $k>K_{\mathrm{cap}}$ remain padded across the training distribution, which helps
prevent previously padded dimensions from activating under higher densities when using running-statistics observation normalization.

To provide global context beyond the fixed neighbor budget, we append a small set of bounded scalars computed from \(\{d_{t,i},\Delta v_t^{i}\}_{i=1}^N\).
In our implementation, these include: (i) a crowd pressure proxy \(P_t\) denoting the log-scaled magnitude of an aggregated repulsive interaction vector,
(ii) its alignment with the ego velocity \(A_t\), (iii) collision-risk statistics characterizing the inverses of the smallest few distances,
(iv) normalized proxemic-zone occupancies \(n_t(r)/K_{\max}\) for several radii \(r\), (v) the active fraction \(N/K_{\max}\), and
(vi) mean relative motion \(\bar{\Delta v}_t\) of nearby pedestrians. Specifically,
\begin{align}
o_t^{\mathrm{sum}}
=
\Big[P_t,A_t,&\{(d_{t,(j)}+\epsilon)^{-1}\}_{j=1}^{J}, \nonumber\\
&\{n_t(r_\ell)/K_{\max}\}_{\ell=1}^{L}, N/K_{\max},\bar{\Delta v}_t\,\Big],
\label{eq:osum_def_compact}
\end{align}
where \(d_{t,(j)}\) is the \(j\)-th smallest distance among \(\{d_{t,i}\}\), and
\(\bar{\Delta v}_t = \frac{1}{\max(1,n_t(r_v))}\sum_{i:\,d_{t,i}\le r_v}(v_t^{i}-v_t^{0})\),.
All components are clipped to fixed ranges for numerical stability, yielding density-comparable summaries that stabilize the policy input under changing \(N\), with $r_v>0$ being a fixed neighborhood radius used for computing velocity statistics. We use standard running-statistics observation normalization and design $ \psi(\cdot)$ so the resulting normalized features remain semantically consistent under varying N.

\subsection{Potential-Based Social Reward Shaping}
\label{subsec:pss}

In dense crowds, collision events are sparse but critical, and large collision penalties can encourage overly conservative behaviors such as freezing.
We therefore add an intrinsic shaping term that provides smooth, anticipatory gradients for socially compliant navigation while remaining stable as density increases. We optimize the return in \eqref{eq:rl_objective} using a reward of the form
\begin{equation}
r_t = r_t^{\mathrm{ext}} + \beta\, r_t^{\mathrm{pss}},
\label{eq:total_reward}
\end{equation}
where $r_t^{\mathrm{ext}}$ is the environment reward including per-step rewards and collision penalties, while
$r_t^{\mathrm{pss}}$ is our Potential-based Social Shaping (PSS) term with weight $\beta\ge 0$.

The proxemic costs are defined as follows, let $d_{t,i}=\|p_t^{i}-p_t^{0}\|$.
We define two proxemic zones: intimate zone and personal zone with corresponding thresholds $d_I<d_P$. We assign distance-only penalties to the two zones as follows:
\begin{align}
C_I(s_t)
&=\sum_{i=1}^{N}\mathbb{I}[d_{t,i}<d_I]\;\phi_I(d_{t,i}), \label{eq:intimate_cost_general}\\
C_P(s_t)
&=\sum_{i=1}^{N}\mathbb{I}[d_I\le d_{t,i}<d_P]\;\phi_P(d_{t,i}), \label{eq:personal_cost_general}
\end{align}
where the per-neighbor penalties are instantiated as distance-only functions:
\begin{align}
\phi_I(d) &= k_{\mathrm{rep}}
\exp\!\left(\operatorname{clip}_{[-c,c]}\!\left(\frac{d_I-d}{\sigma_I}\right)\right),
\;\; d<d_I,
\label{eq:phi_I_def}\\
\phi_P(d) &= \kappa_P\,(d_P-d),
\qquad d_I \le d < d_P,
\label{eq:phi_P_def}
\end{align}
with constants $k_{\mathrm{rep}},\sigma_I,\kappa_P>0$. The clipping operator
$\operatorname{clip}_{[-c,c]}(\cdot)$ is used to prevent numerical overflow in the exponential term.

Since the sums in \eqref{eq:intimate_cost_general}-\eqref{eq:personal_cost_general} can grow with the local interaction load, we rescale them using a
density-adaptive factor based on the local count $n_t(r)$ from \eqref{eq:local_count}. This prevents proxemic penalties from growing simply because more neighbors exist, keeping the policy goal-directed in dense interactions. Specifically, let $r_s$ be a social interaction radius and define
\begin{equation}
\eta_t = \eta\big(n_t(r_s)\big),
\label{eq:density_scale_general}
\end{equation}
where $\eta(\cdot)$ is a non-increasing function that down-weights proxemic costs as the neighborhood becomes crowded.
This prevents the shaping term from dominating the objective solely due to higher density, helping maintain goal-directed behavior in dense interactions.

We further define the potential
\begin{equation}
\Phi(s_t)
=
-\;w_g\|p_t^{0}-g\|
-\;\eta_t\Big(w_I C_I(s_t)+w_P C_P(s_t)\Big),
\label{eq:phi_density_adaptive}
\end{equation}
and convert it into a shaping reward via the standard potential-based form
\begin{equation}
r_t^{\mathrm{pss}}
=
\gamma\,\Phi(s_{t+1})-\Phi(s_t).
\label{eq:pss_reward}
\end{equation}
This yields a dense learning signal that discourages proxemic violations before collisions occur, while the density-adaptive scaling keeps the intrinsic magnitude well-conditioned in crowded scenes. We note that the policy-invariance guarantee of potential-based shaping \cite{ng1999policy} applies to fully observed MDPs. Because our setting is a POMDP and the potential $\Phi(s_t)$ is computed from the full simulator state, we treat the shaping as a training-time learning aid rather than a formal invariance guarantee; the potential uses privileged information available only during training and is not required at deployment.

% Preamble — use EXACTLY ONE of these algorithm packages:
%   \usepackage[ruled,linesnumbered]{algorithm2e}
%
% Remove any of these if present (they conflict):
%   \usepackage{algorithmic}
%   \usepackage{algpseudocode}
%   \usepackage{algorithmicx}

\begin{algorithm}[t]
\caption{PSS-Social}\label{alg:pss_training}
\KwIn{Policy $\pi_\theta$, discount $\gamma$, zone thresholds $d_I < d_P$, PSS weights $w_g, w_I, w_P$, density scaling $\eta(\cdot)$ with radius $r_s$, shaping schedule $\beta_0 \to \beta_T$, slot budget $K_{\max}$ and cap $K_{\mathrm{cap}}$}

Initialize $\pi_\theta$\;
Reset environment; observe $s_0$\;

Compute $d_{0,i} \leftarrow \|p_0^i - p_0^0\|$ for all  $i$\;
$n_0 \leftarrow \sum_i \mathbb{I}[d_{0,i} \le r_s]$; $\eta_0 \leftarrow \eta(n_0)$\;
$C_I \leftarrow \sum_{i} \mathbb{I}[d_{0,i} < d_I]\;\phi_I(d_{0,i})$\;
$C_P \leftarrow \sum_{i} \mathbb{I}[d_I \le d_{0,i} < d_P]\;\phi_P(d_{0,i})$\;
$\Phi_{\mathrm{prev}} \leftarrow -w_g \|p_{0}^0 - g\| - \eta_0 (w_I C_I + w_P C_P)$\;

\For{\textnormal{each training iteration}}{
    Update $\beta$ via linear anneal from $\beta_0$ to $\beta_T$\;

    \For{\textnormal{each environment step} $t$}{

        Compute $d_{t,i} \leftarrow \|p_t^i - p_t^0\|$ for all  $i$\;
        Sort by distance; $K_t \leftarrow \min(N, K_{\mathrm{cap}})$; fill the $K_t$ nearest into $K_{\max}$ slots; pad remaining with $z_{\mathrm{pad}}$\;
        Compute crowd summaries $o_t^{\mathrm{sum}}$\;
        Assemble $o_t \leftarrow [o_t^{\mathrm{ego}};\ o_t^{\mathrm{knn}};\ o_t^{\mathrm{sum}}]$\;

        Sample $a_t \sim \pi_\theta(\cdot \mid o_t)$; execute $a_t$\;
        Observe $s_{t+1}$, $r_t^{\mathrm{ext}}$, and done flag $d_t$\;

        Compute $d_{t+1,i} \leftarrow \|p_{t+1}^i - p_{t+1}^0\|$ for all pedestrians $i$\;
        $n_{t+1} \leftarrow \sum_i \mathbb{I}[d_{t+1,i} \le r_s]$; $\eta_{t+1} \leftarrow \eta(n_{t+1})$\;
        $C_I \leftarrow \sum_{i} \mathbb{I}[d_{t+1,i} < d_I]\;\phi_I(d_{t+1,i})$\;
        $C_P \leftarrow \sum_{i} \mathbb{I}[d_I \le d_{t+1,i} < d_P]\;\phi_P(d_{t+1,i})$\;
        $\Phi_{\mathrm{next}} \leftarrow -w_g \|p_{t+1}^0 - g\| - \eta_{t+1} (w_I C_I + w_P C_P)$\;
        $r_t^{\mathrm{pss}} \leftarrow \gamma\,\Phi_{\mathrm{next}} - \Phi_{\mathrm{prev}}$\;
        $\Phi_{\mathrm{prev}} \leftarrow \Phi_{\mathrm{next}}$\;

        $r_t \leftarrow r_t^{\mathrm{ext}} + \beta\, r_t^{\mathrm{pss}}$\;

        \If{$d_t$}{
            Reset environment; observe $s_0$\;
            Compute $d_{0,i}$, $n_0$, $\eta_0$, $C_I$, $C_P$ as above\;
            $\Phi_{\mathrm{prev}} \leftarrow -w_g \|p_{0}^0 - g\| - \eta_0 (w_I C_I + w_P C_P)$\;
        }
    }

    Update $\theta$ via PPO on collected rollout\;
}
\end{algorithm}

\section{Experimental Results}
\label{sec:results}

We evaluate whether the proposed observation design and reward shaping enable the policy network to maintain collision-free navigation at crowd densities that exceed the training scenarios.
All learning-based methods share an identical PPO training pipeline \cite{schulman2017proximal} and are evaluated on the same density sweep.

% ======================================================================
\subsection{Experimental Setup}
\label{sec:exp_setup}
% ======================================================================

\begin{table}[t]
\centering
\caption{{Safe Success Rate (\%) across the full density sweep.}
Mean $\pm$ 1 std across 5 seeds. $\dagger$: privileged oracle access.}
\label{tab:density_summary}
\vspace{2pt}
\resizebox{\columnwidth}{!}{%
\begin{tabular}{l c c c c c c}
\toprule
\textbf{Method} & $N{=}11$ & $N{=}13$ & $N{=}15$ & $N{=}17$ & $N{=}19$ & $N{=}21$ \\
\midrule
PSS-Social (ours) & $\mathbf{96.2}_{\pm2.2}$ & $\mathbf{95.4}_{\pm1.5}$ & $\mathbf{94.8}_{\pm2.9}$ & $\mathbf{94.0}_{\pm2.5}$ & $\mathbf{93.6}_{\pm2.4}$ & $\mathbf{86.4}_{\pm2.5}$ \\
LSTM-RL            & $91.2_{\pm1.9}$ & $92.8_{\pm1.9}$ & $85.8_{\pm2.8}$ & $83.6_{\pm3.4}$ & $80.8_{\pm3.4}$ & $74.8_{\pm5.0}$ \\
DS-RNN             & $43.6_{\pm9.0}$ & $38.4_{\pm5.5}$ & $36.2_{\pm4.5}$ & $28.2_{\pm6.8}$ & $22.0_{\pm7.8}$ & $12.0_{\pm6.5}$ \\
SARL               & $47.0_{\pm4.3}$ & $39.2_{\pm5.3}$ & $34.0_{\pm3.7}$ & $28.0_{\pm4.4}$ & $16.6_{\pm7.2}$ & $\phantom{0}9.8_{\pm7.0}$ \\
\midrule
ORCA$^\dagger$     & $95.2_{\pm3.0}$ & $93.8_{\pm3.3}$ & $92.8_{\pm2.4}$ & $87.8_{\pm3.3}$ & $83.6_{\pm3.9}$ & $74.8_{\pm4.7}$ \\
SFM$^\dagger$      & $42.6_{\pm5.6}$ & $42.2_{\pm1.3}$ & $40.6_{\pm4.6}$ & $36.4_{\pm4.4}$ & $30.0_{\pm4.1}$ & $29.0_{\pm3.4}$ \\
\bottomrule
\end{tabular}%
}
\end{table}

\begin{figure*}[t]
  \centering
  \includegraphics[width=\linewidth]{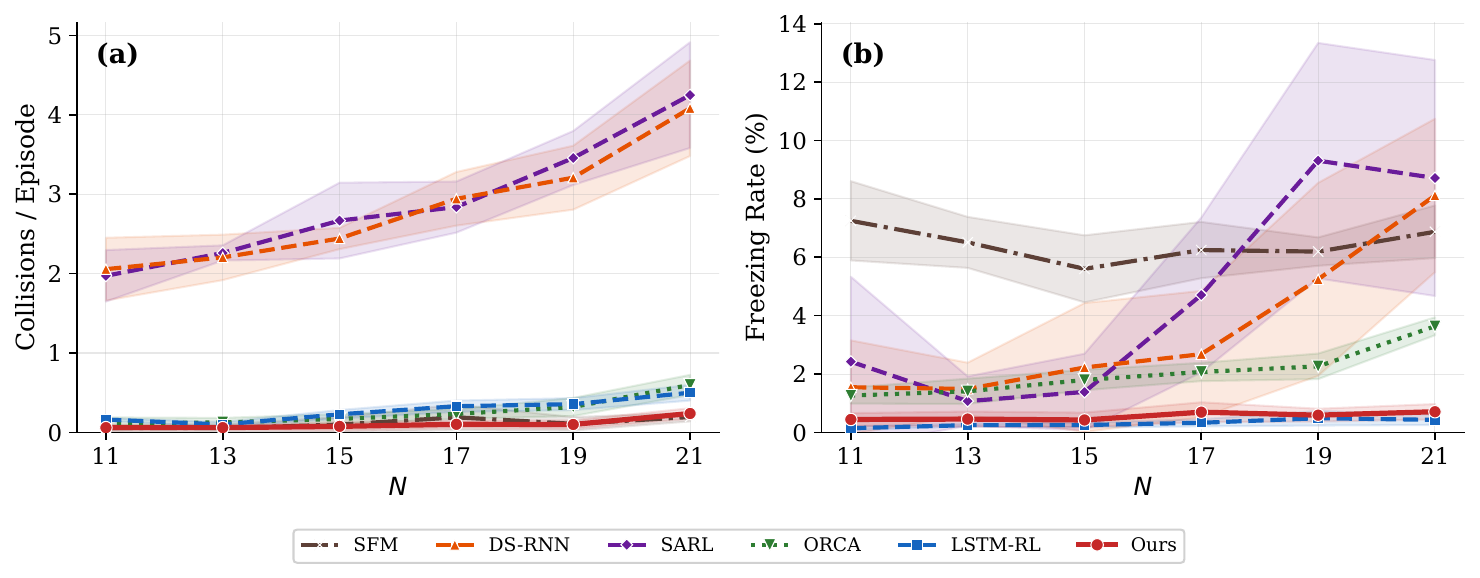}
  \caption{{Analysis of specific failure modes across density.} {(a)}~Collisions per episode measures the frequency of physical contact, distinguishing unsafe behavior from safe navigation. {(b)}~Freezing rate tracks the percentage of time agents spend deadlocked/stationary, identifying conservative failure modes.}
  \label{fig:multipanel}
\end{figure*}

\begin{figure*}[t]
  \centering
  \includegraphics[width=\linewidth]{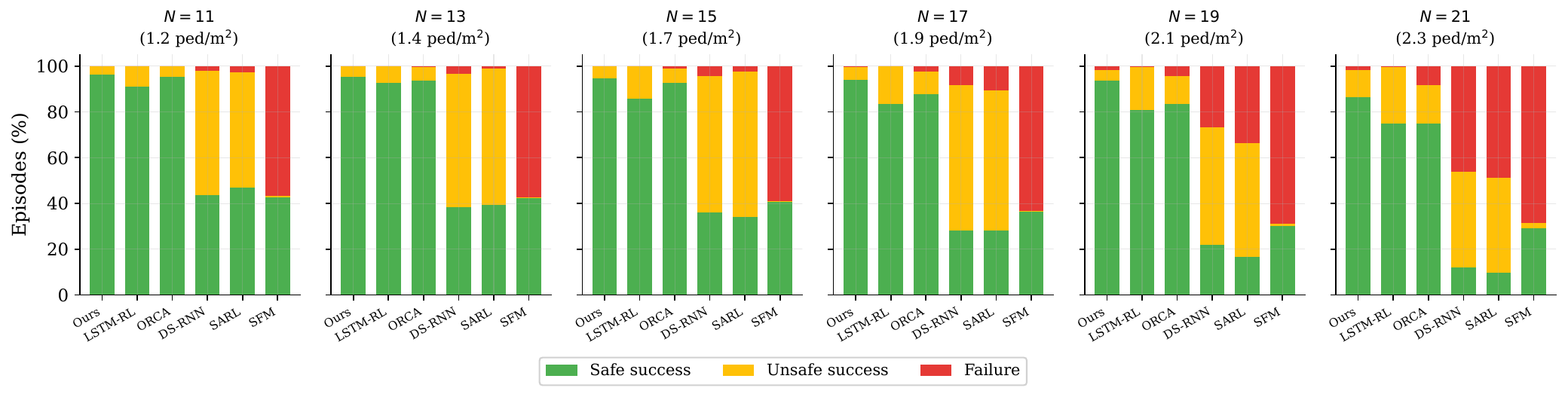}
  \caption{{Episode Outcomes.} Green: Safe success; Amber: Collision; Red: Timeout. Baselines shift from frequent collisions to total failure as density rises, while PSS-Social maintains robust performance.}
  \label{fig:stacked}
\end{figure*}

\textbf{Environment.}
We evaluate in a $3\,\mathrm{m} \times 3\,\mathrm{m}$ arena in which the robot and $N$ pedestrians are initialized at random non-overlapping positions with random goals and a 100-step horizon.
All learning-based agents are trained with density randomization over $N \in [11, 16]$ and tested at six densities $N \in \{11, 13, 15, 17, 19, 21\}$, corresponding to crowd densities $\rho \in [1.22, 2.33]$\,ped/m$^2$.

\textbf{Training.}
All learning-based agents are trained for 12\,M timesteps across 64 parallel environments, with a rollout length of 2048, batch size of 256, and 5 epochs per update.
Training uses a scenario random configuration where the ego and pedestrian's initialization points and the goal location are randomly assigned.

\textbf{Baselines.}
We compare against five established baselines spanning two categories.
Among deep reinforcement learning methods, we evaluate SARL~\cite{chen2019crowd}, DS-RNN~\cite{liu2021decentralized}, and LSTM-RL, which is a recurrent PPO agent implemented to process the same $K$-nearest-neighbor observation used by our method through a two-layer LSTM encoder \cite{raffin2021stable}. We aim to isolate the effect of architectural recurrence from the observation and reward design choices introduced in this work.
As analytic algorithms, we include ORCA~\cite{van2011reciprocal} and a SFM controller \cite{helbing1995social}.
Both analytic methods operate with privileged access to the ground-truth positions and velocities of all pedestrians in the scene, information that is unavailable to any of the learned policies.
They therefore represent oracle upper bounds on what reactive planning can achieve in this environment rather than fair comparisons with the learned methods.
All six methods, including our own, are evaluated with 5 random seeds and 100 episodes per seed at each density level, yielding 500 episodes per condition.

\textbf{Metrics.}
We report three complementary metrics.
(i) Safe Success Rate, which is the primary metric, measures the fraction of episodes in which the robot reaches its goal with zero collisions. We accordingly define unsafe success rate as robot reaches the goal with one or multiple collisions. (ii) Collisions per Episode, which  provides a continuous safety measure. (iii) Freezing Rate, which records the fraction of timesteps in which the robot velocity falls below a low-speed threshold, capturing the pathological deadlock behavior that arises when a policy becomes overly cautious.

% ======================================================================
\subsection{Density Generalization}
\label{sec:main_results}
% ======================================================================

\begin{figure}[t]
  \centering
  \includegraphics[width=\linewidth]{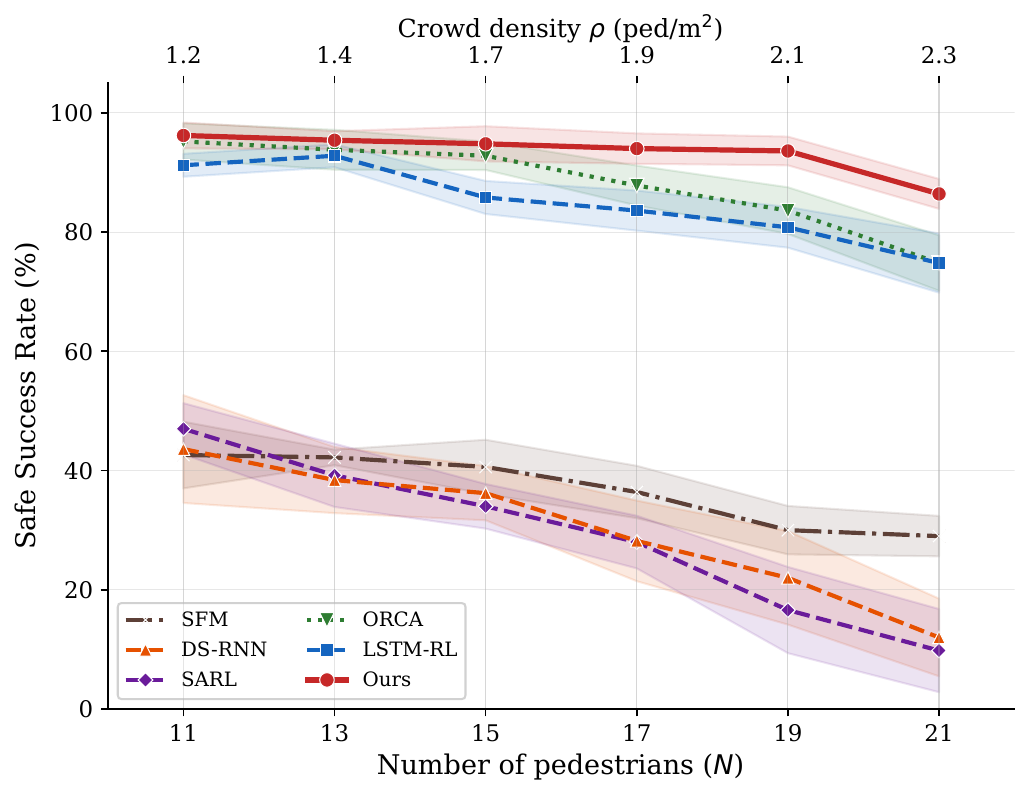}
  \caption{{Safe Success Rate across the density sweep}. }
  \label{fig:hero}
\end{figure}

Fig.~\ref{fig:hero} and Table~\ref{tab:density_summary} present the comparison result of the Safe Success Rates among the algorithms.
PSS-Social achieves a mean Safe Success Rate of ${95.5\%}$ across in-distribution densities ($N \leq 15$) and, critically, sustains ${93.6\%}$ at $N = 19$ and ${86.4\%}$ at $N = 21$, densities that are 19\% and 31\% above the training maximum respectively.
This graceful degradation profile, losing fewer than 10 percentage points from the lightest to the densest test condition, stands in sharp contrast to the behavior of the attention-based DRL baselines.

As illustrated in Fig.~\ref{fig:hero}, DS-RNN and SARL demonstrate limited effectiveness, failing to achieve a majority Safe Success Rate even at the lowest tested density within the training range.
The performance curves for these two methods reveal a degradation as crowd density increases: both exhibit a steep, monotonic decline and retain only a small fraction of their initial performance by $N = 21$ (Table~\ref{tab:density_summary}).
This failure pattern is also reflected in the high collision rates and rising freezing behavior shown in Fig.~\ref{fig:multipanel}.

LSTM-RL, which shares the same observation encoding as PSS-Social but lacks proxemic shaping, provides a controlled comparison that isolates the contribution of the reward design.
As Fig.~\ref{fig:hero} shows, LSTM-RL maintains reasonable performance across the sweep, tracking closer to our method than to the attention-based baselines, yet a widening gap emerges in the out-of-distribution regime. At $N = 17$, $N = 19$, and $N = 21$, PSS-Social consistently outperforms LSTM-RL across seeds. This consistent gap indicates that  proxemic shaping provides a measurable safety benefit beyond what the observation design alone affords, precisely in the density regime where it matters most.

Notably, the oracle planner ORCA, despite operating with privileged access to all agents' true positions and velocities, follows a declining trajectory similar to LSTM-RL and falls significantly below PSS-Social across all out-of-distribution densities with increasing freezing rates shown in Fig.~\ref{fig:multipanel}. This suggests that the proxemic shaping enables qualitatively different avoidance strategies that scale more favorably than geometric reciprocal velocity reasoning in dense crowds.

\begin{table}[t]
\centering
\caption{Reward shaping ablation. Safe Success Rate (\%) across the density sweep.
Mean $\pm$ 1 std across 5 seeds.}
\label{tab:ablation}
\vspace{2pt}
\resizebox{\columnwidth}{!}{%
\begin{tabular}{l c c c c c c}
\toprule
\textbf{Variant} & $N{=}11$ & $N{=}13$ & $N{=}15$ & $N{=}17$ & $N{=}19$ & $N{=}21$ \\
\midrule
Baseline             & $95.8_{\pm1.5}$ & $\mathbf{95.4}_{\pm1.8}$ & $\mathbf{95.6}_{\pm1.7}$ & $92.6_{\pm3.0}$ & $88.0_{\pm4.1}$ & $79.6_{\pm6.7}$ \\
PSS only        & $\mathbf{97.8}_{\pm1.6}$ & $94.2_{\pm3.3}$ & $93.2_{\pm3.0}$ & $91.6_{\pm2.7}$ & $90.8_{\pm3.2}$ & $80.4_{\pm7.4}$ \\
PSS-Social (ours)    & $96.2_{\pm2.2}$ & $95.4_{\pm1.5}$ & $94.8_{\pm2.9}$ & $\mathbf{94.0}_{\pm2.5}$ & $\mathbf{93.6}_{\pm2.4}$ & $\mathbf{86.4}_{\pm2.5}$ \\
\bottomrule
\end{tabular}%
}
\end{table}

\subsection{Ablation Studies}
\label{subsec:ablation}

This section reports two complementary ablation studies that isolate the contributions of (i) reward shaping and (ii) observation encoding. Both studies follow the same training and evaluation protocol in Section~\ref{sec:exp_setup}, specifically, the same density ranges, scenarios, rollout length, and evaluation metrics with only the ablated components are changed. We report Safe Success Rate over a density sweep parameterized by the number of pedestrians $N$, with $N>16$ corresponding to zero-shot density generalization.

\subsubsection{Reward shaping components}
All variants share the same density-invariant observation encoding and identical policy network configuration; they differ only in which shaping components are enabled, with the collision penalty included in the extrinsic reward $r_t^{\mathrm{ext}}$ for all methods. Baseline uses only $r_t^{\mathrm{ext}}$. PSS only adds the proxemic shaping signal but disables density-adaptive scaling, which is equivalent to setting $\eta_t=1$ in \eqref{eq:density_scale_general}. PSS-Social (ours) enables the full PSS with density-adaptive scaling, where the proxemic cost is down-weighted as the local interaction load $n_t(r)$ increases.
At in-distribution densities ($N{=}11,13,15$), all three methods achieve high safe success and the performance gaps are small (Baseline is best at $N{=}13$ and $N{=}15$, while PSS only is best at $N{=}11$). This indicates that in the in-distribution regime, the observation design already supports near-saturated collision-free goal reaching, so shaping yields limited and occasionally mixed improvements.
As density increases beyond the training range, performance gaps widen and PSS-Social is best at all tested zero-shot densities (94.0\% at $N{=}17$, 93.6\% at $N{=}19$, and 86.4\% at $N{=}21$). The gains are most pronounced at high density (e.g., 86.4\% vs.\ 79.6\% for Baseline at $N{=}21$), showing that shaping is most valuable when multi-agent constraints stack and safe success otherwise degrades. Comparing PSS only to PSS-Social isolates the effect of density-adaptive scaling: without scaling, the proxemic shaping magnitude can grow with the number of nearby pedestrians, making the intrinsic term less well-conditioned as $n_t(r)$ increases; density-adaptive scaling counteracts this effect and yields stronger zero-shot performance (e.g., 86.4\% vs.\ 80.4\% at $N{=}21$), along with improved robustness across seeds as reflected by reduced variance in the hardest settings.

\subsubsection{Observation encoding design} 
We next ablate the two ingredients of the density-invariant observation encoding using a $2{\times}2$ design while keeping the policy network and reward shaping fixed to PSS-Social. The first factor is $K$-cap truncation: whether we populate only the closest $K$ slots and keep remaining slots at a constant padding value. The second factor is distance sorting: whether each slot corresponds to the $k$-th closest pedestrian or uses an unsorted assignment.
Table~\ref{tab:kcap_ablation} shows that distance sorting is necessary for stable learning: removing sorting causes a large drop in safe success even at low density due to unstable slot semantics. $K$-cap truncation is critical for zero-shot density generalization: with sorting enabled, removing $K$-cap has little effect for $N\le 15$ but leads to a sharp collapse for $N>16$, consistent with the mechanism that previously padded feature dimensions begin to activate under higher densities and destabilize the normalized input distribution. Together, these results support the observation-side contribution: distance sorting stabilizes per-slot meaning, and $K$-cap truncation prevents distribution shift at the padding boundary when evaluating beyond the training crowd size.

\begin{table}[t]
\centering
\caption{Observation encoding ablation. Safe Success Rate (\%) across the density sweep.
Mean $\pm$ 1 std across 5 seeds.}
\label{tab:kcap_ablation}
\vspace{2pt}
\resizebox{\columnwidth}{!}{%
\begin{tabular}{l c c c c c c}
\toprule
\textbf{Variant} & $N{=}11$ & $N{=}13$ & $N{=}15$ & $N{=}17$ & $N{=}19$ & $N{=}21$ \\
\midrule
$K$-cap + Sort (full)      & $96.2_{\pm 2.2}$ & $95.4_{\pm 1.5}$ & $94.8_{\pm 2.9}$ & $94.0_{\pm 2.5}$ & $93.6_{\pm 2.4}$ & $86.4_{\pm 2.5}$  \\
No $K$-cap + Sort          & $96.2_{\pm 2.2}$ & $95.4_{\pm 1.5}$ & $94.8_{\pm 2.9}$ & $25.4_{\pm 20.8}$ & $5.6_{\pm 3.4}$  & $2.0_{\pm 1.6}$  \\
$K$-cap + No Sort          & $39.0_{\pm 5.4}$ & $32.0_{\pm 2.4}$ & $27.0_{\pm 5.8}$ & $24.4_{\pm 2.9}$ & $19.2_{\pm 6.3}$ & $14.8_{\pm 2.5}$  \\
No $K$-cap + No Sort       & $39.0_{\pm 5.4}$ & $32.0_{\pm 2.4}$ & $27.0_{\pm 5.8}$ & $7.0_{\pm 4.7}$  & $2.8_{\pm 2.2}$  & $1.0_{\pm 1.4}$  \\
\bottomrule
\end{tabular}%
}
\end{table}
% --------------------------------------------------------------------------
% ======================================================================
\subsection{Summary of Findings}
\label{sec:results_summary}
% ======================================================================

The experiments support three main conclusions.
First, the combination of KNN-sorted, fixed-length observations with density-invariant normalization enables a simple MLP policy to generalize to crowd densities 31\% beyond its training distribution, achieving 86.4\% collision-free success at $N = 21$ where state-of-the-art attention-based methods retain only a fraction of their in-distribution performance.
Second, potential-based social shaping with density-adaptive normalization provides a statistically significant safety margin over the same observation encoder trained without shaping, across the critical out-of-distribution density range from $N = 17$ to $N = 21$.
Third, the proposed method avoids the conservative deadlock that afflicts reactive social force planners, maintaining freezing below 1\% and goal-reaching above 98\% at all tested conditions.
Together, these results validate the central thesis that {observation design} and {reward design}, rather than architectural complexity, are the limiting factors for density generalization in DRL-based crowd navigation.

\section{Conclusion}
We studied dense crowd navigation with {zero-shot density generalization}, where the number of pedestrians at deployment exceeds the training range. We proposed  a reinforcement learning framework that improves robustness to density shifts through two complementary designs: a {density-invariant observation encoding} that preserves stable input semantics via distance-sorted KNN slots augmented with bounded crowd-summary scalars, and {potential-based proxemic reward shaping} with {density-adaptive scaling} to keep the intrinsic social signal well-conditioned as local interaction load increases.

Across a controlled density sweep, our method maintains strong collision-free performance beyond the training distribution, achieving high safe success in the out-of-distribution regime while avoiding conservative deadlock behaviors reflected by low freezing rates. Ablations further confirm that both design choices are beneficial. % The $K$-cap, distance-sorted observation encoding preserves stable input semantics under density shifts, and the potential-based proxemic shaping with density-adaptive scaling provides an additional safety margin and improved robustness in the zero-shot high-density regime.

\addtolength{\textheight}{-12cm}   % This command serves to balance the column lengths
                                  % on the last page of the document manually. It shortens
                                  % the textheight of the last page by a suitable amount.
                                  % This command does not take effect until the next page
                                  % so it should come on the page before the last. Make
                                  % sure that you do not shorten the textheight too much.

%%%%%%%%%%%%%%%%%%%%%%%%%%%%%%%%%%%%%%%%%%%%%%%%%%%%%%%%%%%%%%%%%%%%%%%%%%%%%%%%

%%%%%%%%%%%%%%%%%%%%%%%%%%%%%%%%%%%%%%%%%%%%%%%%%%%%%%%%%%%%%%%%%%%%%%%%%%%%%%%%

%%%%%%%%%%%%%%%%%%%%%%%%%%%%%%%%%%%%%%%%%%%%%%%%%%%%%%%%%%%%%%%%%%%%%%%%%%%%%%%%
% \section*{APPENDIX}

% Here is a placeholder for appendix

%%%%%%%%%%%%%%%%%%%%%%%%%%%%%%%%%%%%%%%%%%%%%%%%%%%%%%%%%%%%%%%%%%%%%%%%%%%%%%%%

\bibliographystyle{IEEEtran}
\bibliography{main}
\end{document}